\documentclass[conference]{IEEEtran}
\IEEEoverridecommandlockouts
\renewcommand{\thesubsection}{\thesection.\alph{subsection}}
\usepackage{cite}
\usepackage{amsmath,amssymb,amsfonts}
\usepackage{algorithmic}
\usepackage{graphicx}
\usepackage{textcomp}
\usepackage{xcolor}
\usepackage{enumitem}
\usepackage{float}
\usepackage{booktabs}
\usepackage{graphicx}
\usepackage{fancyhdr}
\usepackage{subcaption}
\usepackage{caption}
\usepackage[multiple]{footmisc}
\usepackage{hyperref}
\hypersetup{
    colorlinks=true,
    linkcolor=blue,
    filecolor=magenta,      
    urlcolor=magenta,
}
\ifCLASSINFOpdf
\else
\fi
\begin{document}
\title{Improved Actor Relation Graph based Group Activity Recognition}

\author{\IEEEauthorblockN{Zijian Kuang}
\IEEEauthorblockA{Department of Computing Science\\
University of Alberta\\
Edmonton, Canada \\
Email: kuang@ualberta.ca}
\and
\IEEEauthorblockN{Xinran Tie}
\IEEEauthorblockA{Department of Computing Science\\
University of Alberta\\
Edmonton, Canada \\
Email: xtie@ualberta.ca}
}


%Telephone: (800) 555--1212, Fax: (888) 555--1212}


\maketitle

\begin{abstract}
Video understanding is to recognize and classify different actions or activities appearing in the video. A lot of previous work, such as video captioning, has shown promising performance in producing general video understanding. However, it is still challenging to generate a fine-grained description of human actions and their interactions using state-of-the-art video captioning techniques. The detailed description of human actions and group activities is essential information, which can be used in real-time CCTV video surveillance, health care, sports video analysis, etc. This study proposes a video understanding method that mainly focused on group activity recognition by learning the pair-wise actor appearance similarity and actor positions. We propose to use Normalized cross-correlation (NCC) and the sum of absolute differences (SAD) to calculate the pair-wise appearance similarity and build the actor relationship graph to allow the graph convolution network to learn how to classify group activities. We also propose to use MobileNet as the backbone to extract features from each video frame. A visualization model is further introduced to visualize each input video frame with predicted bounding boxes on each human object and predict individual action and collective activity.
\end{abstract}


% \fi
\IEEEpeerreviewmaketitle

\section{Introduction}
Video understanding is an extensively studied topic widely used in the video content analysis area \cite{Video}. Traditional video captioning techniques such as LSTM-YT \cite{Translating} and S2VT \cite{seq2seq} use recurrent neural networks, specifically LSTMs \cite{fundamentals_of_RNN}, to train the models with video-sentence pairs \cite{seq2seq, dense_captioning, Video, Translating}. The models can learn the association between video frames' sequence and the sequence of sentences to generate a description of videos \cite{seq2seq}. Krishna et al. indicated that those video captioning approaches only works for a short video with only one major event \cite{dense_captioning}. Therefore, they introduced a new captioning module that uses contextual information from the timeline to describe all the events during a video clip \cite{dense_captioning}. However, it is still very limited in video captioning approaches to generate a detailed description of human actions, and their interactions \cite{Weakly}. Recent studies in pose estimation, human action recognition, and group activity recognition areas show the capability to describe more detailed human actions, and human group activities \cite{har, openpose}.

Human action recognition and group activity recognition are an important problem in video understanding \cite{arg}. The action and activity recognition techniques have been widely applied in different areas such as social behavior understanding, sports video analysis, and video surveillance. To better understanding a video scene that includes multiple persons, it is essential to understand both each individual’s action and their collective activity. Actor Relation Graph (ARG) based group activity recognition is the state-of-the-art model that focuses on capturing the appearance and position relationship between each actor in the scene and performing the action and group activity recognition \cite{arg}. 

In this paper, we propose several approaches to improve the functionality and the performance of the Actor Relation Graph-based model to perform a better video understanding that mainly focused on group activity recognition. To enhance human action and group activity recognition performance, we apply MobileNet in the CNN layer and use Normalized cross-correlation (NCC) and the sum of absolute differences (SAD) to calculate the pair-wise appearance similarity to build the Actor Relation Graph. We also introduce a visualization model that plots each input video frame with predicted bounding boxes on each human object and predicted individual action and group activity. The output examples are shown in Fig. \ref{fig:3}.\

\section{Related Work}
\label{RW}
\subsection{Video Captioning}
One important study area of video understanding is video captioning. In 2015, S. Venugopalan et al. proposed an end-to-end sequence-to-sequence model which exploited recurrent neural network, specifically Long Short-Term Memory (LSTM \cite{fundamentals_of_RNN}) networks as trained on video-sentence pairs and learned to associate a sequence of frames in a video to sequential words to generate the descriptions of the event in the video as captions \cite{seq2seq}. A stack of two LSTMs was used to learn the frames' sequence's temporal structure and the sequence model of the generated sentences. In this approach, the entire video sequence needs to be encoded using the LSTM network at the beginning. Long video sequences could lead to vanishing gradients and prevent the model from being trained successfully \cite{dense_captioning}.

In 2017, R. Krishna et al. introduced a Dense-Captioning Events (DCE) model that can detect multiple events and generate a description for each event using the contextual information from past, concurrent, and future in a single pass of the video \cite{dense_captioning}. In this paper, the process is divided into two steps: event detection and description of detected events. The DCE model leverages a multi-scale variant of the deep action proposal model to localize temporal proposals of interest in short and long video sequences. In addition, a captioning LSTM model is introduced to exploit the context from the past and future with an attention mechanism.

In 2018, X. P. Li et al. introduced a novel attention-based framework called Residual attention-based LSTM (Res-ATT \cite{residual_attention_LSTM}). This new model benefits from the existing attention mechanism and further integrates the residual mapping into a two-layer LSTM network to avoid losing previously generated words information. The residual attention-based decoder model is designed with five separate parts: a sentence encoder, temporal attention, a visual and sentence feature fusion layer, a residual layer, and an MLP \cite{residual_attention_LSTM}. The sentence encoder is an LSTM layer that explores important syntactic information from a sentence. The temporal attention is designed to identify the importance of each frame. The visual and sentence feature fusion LSTM layer is working on mixing natural language information with image features, and the residual layer is proposed to reduce the transmission loss. The MLP layer is used to predict the word to generate a description in natural language \cite{residual_attention_LSTM}.

\subsection{Real-time Pose Estimation and Human Activity Recognition}
To better understanding a video scene that includes multiple persons, it is essential to understand both each individual’s action. OpenPose is an open-source real-time system which is used for 2D multi-person pose detection \cite{openpose}. Nowadays, it is also widely used in body and facial landmark points detection in video frames \cite{PRT, har, efficient_online}. It produces a spatial encoding of pairwise relationships between body parts for a variable number of people, followed by a greedy bipartite graph matching to output the 2D keypoints for all people in the image.  In this approach, both prediction of part affinity fields (PAFs) and detection of confidence maps are refined at each stage \cite{openpose}. By doing this, the real-time performance is improved while it maintains the accuracy of each component separately. The online OpenPose library supports jointly detect the human body, hand, and facial keypoints on a single image, which provides 2D human pose estimation for our proposed system.

Recurrent Neural Networks (RNNs) with Long Short-Term Memory (LSTM) cells are widely used for human action recognition with the emerging accessible human activity recognition methods. In this paper, F. M. Noori et al. proposes an approach that first extracts anatomical keypoints from RGB images using the OpenPose library and then obtains extra-temporal motions features after considering the movements in consecutive video frames, and lastly classifies the features into associated activities using RNN with LSTM \cite{har}. Improved performance is shown as organizing activities performed by several different subjects from various camera angles. However, their work based on multi-person action classification is still in progress for accuracy improvement.

\subsection{Learning Actor Relation Graphs for Group Activity Recognition}
Human action recognition and group activity recognition are an important problem in video understanding \cite{arg}. In 2019, J. Wu et al. proposed to use Actor Relational Graph (ARG) to model relationships between actors and recognize group activity with multiple persons involved \cite{arg}. Using the ARG in a multi-person scene, the relation between actors from respect to appearance similarity and the relative location is inferred and captured. Compared with using a CNN to extract person-level features and later aggregate the features into a scene-level feature or using RNN to capture temporal information in the densely sampled frames, learning with ARG is less computationally expensive and more flexible while dealing with variation in the group activity. Given a video sequence with bounding boxes and ground truth labels of action for actors in the scene, the trained network can recognize individual actions and group activity in a multi-person scene. For long-range video clips, ARG's efficiency is improved by forcing a relational connection only in a local neighborhood and randomly dropping several frames while maintaining the training samples' diversity and reducing the risk of overfitting. At the beginning of the training process, the actors' features are first extracted by CNN and RoIAlign model\cite{roialign} using the provided bounding boxes. After obtaining the feature vectors for actors in the scene, multiple actor relation graphs are built to represent the diverse information for the same set of actors’ features. Finally, Graph Convolutional Network (GCN) is applied to perform learning and inference to recognize individual actions and group activity based on ARG. Two classifiers used for individual actions and group activity recognition are applied respectively to the pooled ARG. Scene-level representation is generated by max-pooling individual actor representations, which later uses for group activity classification.

\section{Proposed Method}
\label{PM}
\begin{figure}[H]
\centerline{\includegraphics[width=3.5in]{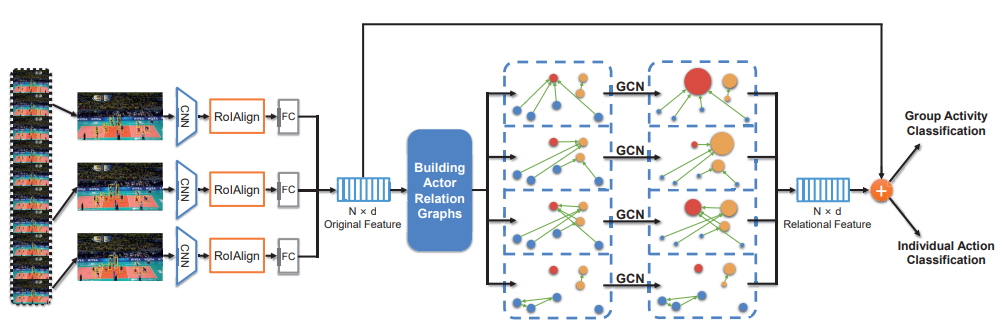}}
\caption{The original ARG-based group activity recognition framework. \cite{arg}}
\label{fig1}
\end{figure}

We propose using an improved Actor relation graph-based model that focused on group activity recognition. The overview of the original ARG-based model is shown in Fig.~\ref{fig1}.

The model first extracts actor features from sampled video frames with manually labeled bounding boxes using CNN, and RoIAlign \cite{roialign}. Next, it builds an N by d dimensional feature matrix, using a d-dimension vector to represent each actor's bounding box and using N to present the total number of bounding boxes in video frames. The actor relation graphs are then built to capture each actor's appearance and position relationship in the scene. Afterward, the model uses Graph Convolutional Networks (GCN) to analyze each actor's relationship from the ARG. Finally, the original and relational features are aggregated and used by two separate classifiers to perform actions, and group activity recognition  \cite{arg}. Since the study is mainly focused on group activity recognition, the individual action recognition is not very accurate because the model only uses the region of interest and CNN to perform action recognition.

\begin{figure}[H]
\centerline{\includegraphics[width=3.6in]{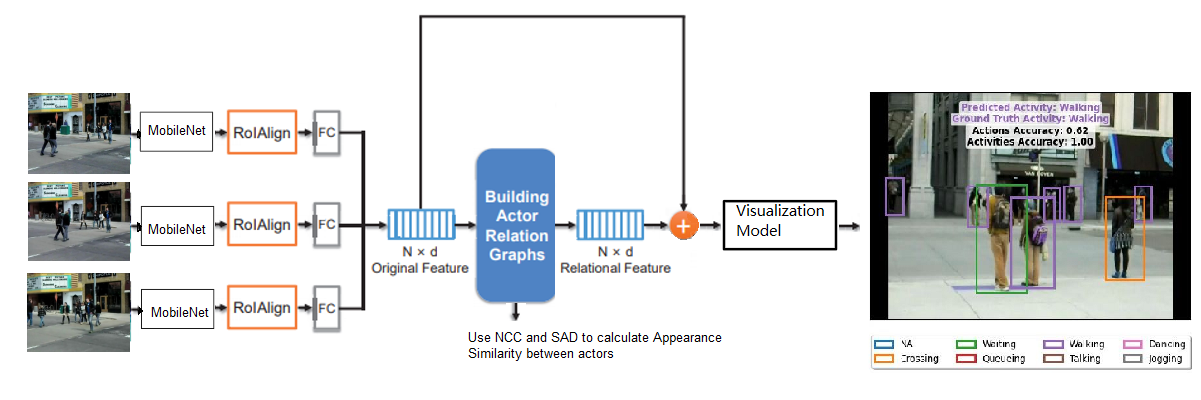}}
\caption{The improved ARG-based human actions and group activity recognition model proposed by us}
\label{fig2}
\end{figure}

Although using the ARG-based model archives high accuracy predictions on group activities. There are still some potential improvement areas. The improved ARG-based human actions and group activity recognition model proposed by us is shown in Fig.~\ref{fig2}.

We propose to apply MobileNet in the CNN layer to extract image feature maps and use Normalized cross-correlation (NCC) and the sum of absolute differences (SAD) to calculate the pair-wise appearance similarity to build the Actor Relation Graph. More details of our proposed methodology are mentioned in section V.

To make our model have a more visualized result, we also introduce a visualization model that plots each input video frame with predicted bounding boxes on each human object and predicted individual action and group activity as the output. The output examples are shown in Fig.  \ref{fig:3}.

\section{Methodology}
\label{methodology}

In our model, how to build this Actor relation graph is the key. J. Wu et al. has proved that the ARG can represent the graph structure of pair-wise relation information between each pair of actors in each frame and use the related information for group activity understanding \cite{arg}.

To better understand the relationship between two actors, both appearance features and position information are used to construct the ARG. The relation value is defined as a composite function below, which function fa indicates the appearance relation, and function fs indicates the position relation. The $x^a_i$ and $x^a_j$ refers to the actor i's and actor j's appearance features, while $x^s_i$ and $x^s_j$ refers to the actor i's and actor j's location features (the center coordinates of each actor's bounding box). The function h fuses appearance and position relation to a scalar weight \cite{arg}:

\begin{equation}
G_{ij} = h(f_a(x^a_i, x^a_j ), f_s(x^s_i, x^s_j)) \label{eq1}
\end{equation}

The normalization is further adopted on each actor node with SoftMax function so that the sum of all the corresponding values of each actor node will always equal to one \cite{arg}:

\begin{equation}
\mathbf{G}_{i j}=\frac{f_{s}\left(\mathbf{x}_{i}^{s}, \mathbf{x}_{j}^{s}\right) \exp \left(f_{a}\left(\mathbf{x}_{i}^{a}, \mathbf{x}_{j}^{a}\right)\right)}{\sum_{j=1}^{N} f_{s}\left(\mathbf{x}_{i}^{s}, \mathbf{x}_{j}^{s}\right) \exp \left(f_{a}\left(\mathbf{x}_{i}^{a}, \mathbf{x}_{j}^{a}\right)\right)}
\end{equation}

\subsection{Appearance relation}
In J. Wu’s paper, the Embedded Dot-Product is implemented to compute the similarity between the two actors' appearance features (the image feature inside each actor’s bounding box) in embedding space \cite{arg}:. The corresponding function is written in the way below:

\begin{equation}
f_{a}\left(\mathbf{x}_{i}^{a}, \mathbf{x}_{j}^{a}\right)=\frac{\theta\left(\mathbf{x}_{i}^{a}\right)^{\mathrm{T}} \phi\left(\mathbf{x}_{j}^{a}\right)}{\sqrt{d_{k}}}
\end{equation}

The $\theta$ and $\phi$ are two functions using Wx + b, in which W and b are learnable weight. The learnable transformations of original appearance features can better understand the relation between two actors in a subspace.

Our model evaluates two other methods for the appearance relation calculation: the Normalized cross-correlation (NCC) and the sum of absolute differences (SAD).

Normalized cross-correlation (NCC) is a method to evaluate the degree of similarity between two compared images. The brightness of the compared images can vary due to lighting and exposure conditions, so the images can be first normalized to get a more accurate similarity score by applying NCC. The advantage of the normalized cross-correlation is that it is less sensitive to linear changes in the amplitude of illumination in the two compared images, and the corresponding function can be written in the following way \cite{NCC}:

\begin{equation}
\varphi_{{x}_{i}^{a} {x}_{j}^{a}}^{\prime}(t)=\frac{\varphi_{{x}_{i}^{a} {x}_{j}^{a}}(t)}{\sqrt{\varphi_{{x}_{i}^{a} {x}_{i}^{a}}(0) \varphi_{{x}_{j}^{a} {x}_{j}^{a}}(0)}}
\end{equation}

The normalized quantity $\varphi_{{x}_{i}^{a} {x}_{j}^{a}}^{\prime}(t)$ will be vary between -1 and 1, while value 1 indicates exactly matching between two images, and value of 0 indicates no matching between two images. The NCC value can help us better understand the appearance relation between each pair of actors.

Sum of absolute differences (SAD) is another method we evaluate to calculate the appearance relation when building the ARG. SAD calculates the distance between two matrices by computing the sum of absolute difference of the components of the matrices as the formula:

\begin{equation}
\operatorname{SAD}\left({x}_{i}^{a}, {x}_{j}^{a}\right)=\sum_{k}^{n}\left|{x}_{ik}^{a}-{x}_{jk}^{a}\right|
\end{equation}

Since SAD is more resistant to extreme values in the data, it is more robust when comparing appearance features and better captures the appearance relation.

\subsection{Position relation}
Besides, spatial structural information is considered to capture the position relation between actors better. A distance mask has been applied to obtain signals from entities that are not distantly apart. Since relation in a local scope is more crucial compared with global relation for group activity understanding, a measure of Euclidean distance $\mathbf{G}_{i j}$ between two actors is computed as:
 
 \begin{equation}
 f_{s}\left(\mathbf{x}_{i}^{s}, \mathbf{x}_{j}^{s}\right)=\mathbb{I}\left(d\left(\mathbf{x}_{i}^{s}, \mathbf{x}_{j}^{s}\right) \leq \mu\right),
 \end{equation}
 
where $\mathbb{I}(\cdot)$ denotes the indicator function, $d(\mathbf{x}_{i}^{s}, \mathbf{x}_{j}^{s})$ calculates the Euclidean distance between the center coordinates of two actors’ bounding boxes, and $\mu$ is a distance threshold.

%Be sure that the symbols in your equation have been defined before or immediately following the equations. Use Eq.~\eqref{eq} to refer to the equations. 

\section{Experiments and Results}
\label{ER}

\subsection{Datasets}
In this paper, we used the public group activity recognition datasets called collective activity dataset and augmented dataset \cite{collective_activity_dataset} to train and test our model. This dataset has 74 video scene that includes multiple persons in each scene. The manually defined bounding boxes on each person and the ground truth of their actions and the group activity are also labeled in each frame.

\subsection{Implementation details and results}
We use a minibatch size of 16 with a learning rate of 0.0001 and train our network in 100 epochs. The individual action loss weight $lambda$ = 1 is used. The GCN parameters are set as dk = 256, ds = 32, and 1/5 of the image width is adopted to be the distance mask threshold µ. The default backbone CNN network for feature extraction is set as Inception-v3, and the default of the appearance relation function is set as embedded dot-product. Our implementation is based on the PyTorch framework and two pieces of RTX 2080 Ti GPUs. \\

\subsubsection{Experiment 1 - evaluate with different backbone networks}
In this subsection, we conduct detailed studies on the Collective Activity dataset to understand the proposed backbone networks' relation modeling using group activity prediction accuracy as the assessment metric. The results of the experiments are shown in Table \ref{tab:backbone_table}. 

In our 2-stage training, we first finetune the ImageNet pre-trained backbone network with the randomly selected frame from each of the training samples. Then, the weights of the feature extraction part of the backbone network are fixed in stage 2. We further train the network with GCN and calculate appearance relation using embedded dot-product. We begin our experiments with Inception-v3 as the backbone network. The first stage of training takes approximately 2.6 hours, while the second stage takes a longer time, about 3.3 hours. With Inception-v3, the group activity recognition accuracy after the training of stage 1 achieves 90.91\%. With the additional training of GCN in stage 2, our model yields a higher recognition accuracy of 92.71\%. 

We further adopt MobileNet as the backbone network to boost the speed of our model. MobileNet is a deep convolutional neural network that is lightweight but efficient. With MobileNet, the training time of stage 1 is 1.8 hours, which reduces the time spent at stage 1 by 32.7\%. The training time of stage 2 is 2.4 hours, which is 26\% less than the training time of Inception-v3 at the same stage. In summary, the training speed of our model is increased by 35\%. However, the activity recognition accuracy is slightly dropped from 92.71\% to 91.44\%.

\begin{table}[htbp]
\caption{Accuracy (\%) of Group Activity Prediction from Two Backbone Networks}
\label{tab:backbone_table}
\resizebox{\columnwidth}{!}{%
\begin{tabular}{l c c p{1.5cm}}
\toprule
Backbone Network & Stage & Time Cost (seconds) & Best Group Activity Prediction Accuracy \\ 
\midrule
Inception-v3     & 1     & 9569.8              & 90.91\%                                 \\
Inception-v3     & 2     & 11784.4             & 92.71\%                                 \\
MobileNet(our model)        & 1     & \textbf{6443.8}              & 89.37\%                                 \\
MobileNet(our model)        & 2     & \textbf{8719.7}              & 91.44\%                                 \\
\bottomrule
\end{tabular}%
}
\end{table}

\subsubsection{Experiment 2 - evaluate with different appearance relation functions}
In this experiment, we evaluate the group activity recognition performance with different appearance relation functions. We first train and validate the group activity recognition performance based on the default Inception-v3 backbone and the embedded dot-product for appearance relation calculation. The best result we get is 92.71\%. Then we update the appearance relation function with Normalized cross-correlation (NCC) to draw the actor relation graph, and the best result we achieve is 93.50\%. We further evaluate the sum of absolute distance (SAD) function to calculate the appearance similarity, and the best score we achieve is 93.98\%. In this experiment 2, we prove that our proposed model with either NCC or SAD as the appearance relation calculation function will achieve better group activity prediction accuracy as expected. The results of the experiments are shown in table \ref{tab:inv3_table}.

\begin{table}[htbp]
\caption{Accuracy (\%) of Group Activity Prediction on Backbone Network Inception-v3 with Different Appearance Relation Functions}
\label{tab:inv3_table}
\centering
\begin{tabular}{@{}lc@{}}
\toprule
Method               & Best Group Activity Prediction Accuracy \\
\midrule
Embedded Dot-product & 92.71\%                                 \\
NCC(our model)                  & \textbf{93.50\%}             \\
SAD(our model)                  & \textbf{93.98\%}             \\   
\bottomrule
\end{tabular}%
\end{table}

\section{Conclusion and Future Work}
This paper utilizes the actor relation graph (ARG) based model with novel improvements for group activity recognition. To enhance our model's performance, we learn ARG to perform appearance relation reasoning on graphs using normalized cross-correlation (NCC) and the sum of absolute difference (SAD). Besides, to improve the computational speed, we introduce MobileNet as the backbone network into our proposed model. Furthermore, extensive experiments demonstrate that the proposed methods are robust and effective for enhancing both accuracy and speed on the Collective Activity dataset. Since our project is mainly focused on group activity recognition, the individual action recognition is not very accurate because we only use the region of interest and CNN to perform action recognition. In the future, we would like to finetune the backbone network MobieNet further and apply skeleton extraction to achieve higher prediction accuracy for individual actions and group activities.

\section*{Acknowledgment}

The authors would like to thank our mentor Dr. Nasim Hajari, Postdoctoral Fellow, Department of Computing Science, University of Alberta, for her guidance and feedback throughout the research and study. We would also thank our advisors Dr. Lihang Ying and Dr. Anup Basu for their motivation and support to bring out the novelty in our research. Finally, We would like to thank the researchers of the previous work, which is the inspiration and starting point for our research.




%\bibliographystyle{IEEEtran}


% \end{thebibliography}

\bibliographystyle{IEEEtran}
\bibliography{main}
\vspace{-1 cm}
\begin{IEEEbiographynophoto}{Zijian Kuang}
is a machine learning enthusiast who is currently pursuing his MSc. in Computing Science at the University of Alberta. He has 5+ years of work experience in software development and project management with experience working in several applications for government and large organizations.
\end{IEEEbiographynophoto}

\vspace{-1 cm}
\begin{IEEEbiographynophoto}{Xinran Tie}
is currently a student pursuing MSc. in Computing Science with Specialization in Multimedia at the University of Alberta. She is expected to graduate in May 2022, and currently studies on and researches 3 projects from the courses. 
\end{IEEEbiographynophoto}

\section*{Contributions}
\begin{table}[H]
\centering
\begin{tabular}{|l|l|}
\hline
Idea &
  \begin{tabular}[c]{@{}l@{}}Existing work research: all members\\ Experiments on novel combinations: all members\\ Experiments on novel techniques: all members\end{tabular} \\ \hline
Coding &
  \begin{tabular}[c]{@{}l@{}}Configuration: Zijian Kuang\\ Implementation of predict class: Zijian Kuang\\ ReadMe file: Zijian Kuang\\ Implementation of NCC and training: Zijian Kuang\\ Implementation of SAD and training: Xinran Tie\\ Change of backbone network and training: Xinran Tie\\ Visualization model: Xinran Tie\\ Building ARG-based Network and training: all members\end{tabular} \\ \hline
Writing &
  \begin{tabular}[c]{@{}l@{}}Abstract and introduction: Zijian Kuang\\ Related Work: all members\\ Proposed Method: Zijian Kuang\\ Methodology: all members\\ Experiments and results: all members\\ Conclusion and acknowledgment: Xinran Tie\\ References: all members\end{tabular} \\ \hline
\end{tabular}
\end{table}

\clearpage
\onecolumn
\appendix

\begin{figure*}[htp!]
    \centering
    \begin{subfigure}[b]{0.49\columnwidth}
        \centering
        \includegraphics[width=\columnwidth]{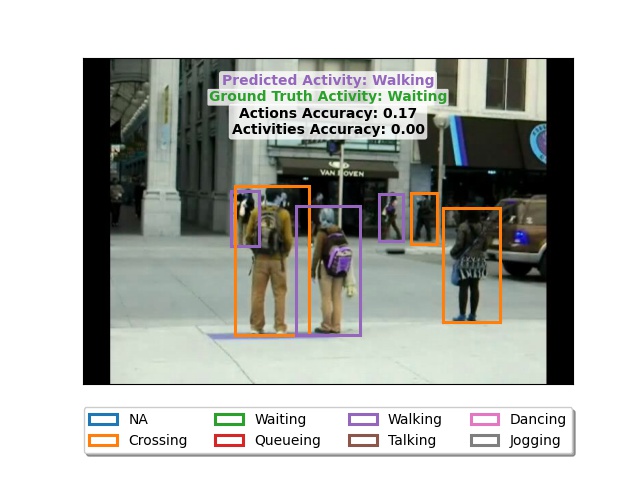}
        \label{fig:1}
    \end{subfigure}
    \begin{subfigure}[b]{0.49\columnwidth}  
        \centering 
        \includegraphics[width=\columnwidth]{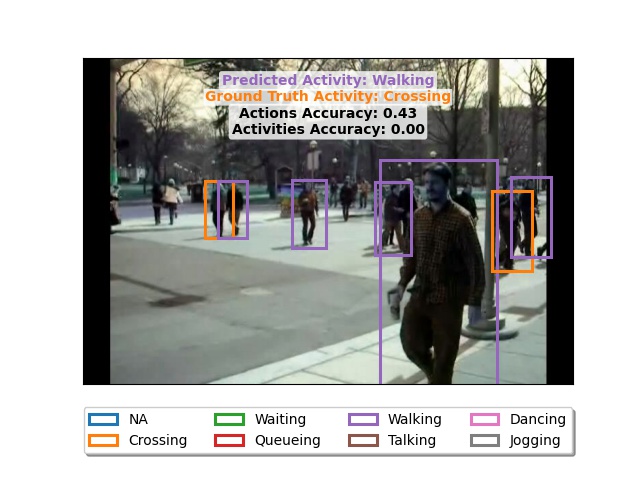}
        \label{fig:2}
    \end{subfigure}
    
     \begin{subfigure}[b]{0.49\columnwidth}  
        \centering 
        \includegraphics[width=\columnwidth]{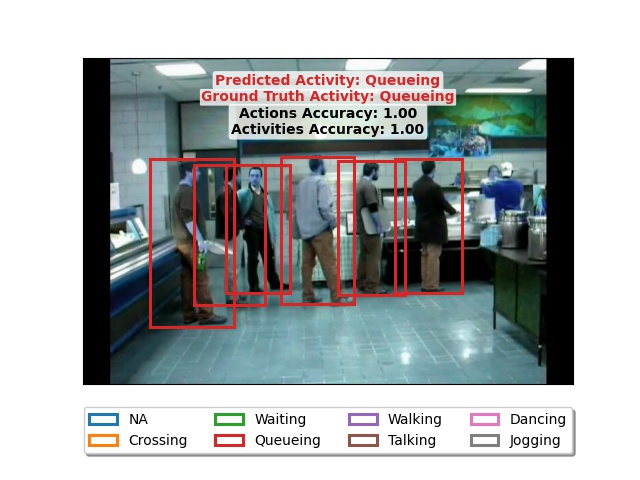}
        \label{fig:2}
    \end{subfigure}
     \begin{subfigure}[b]{0.49\columnwidth}  
        \centering 
        \includegraphics[width=\columnwidth]{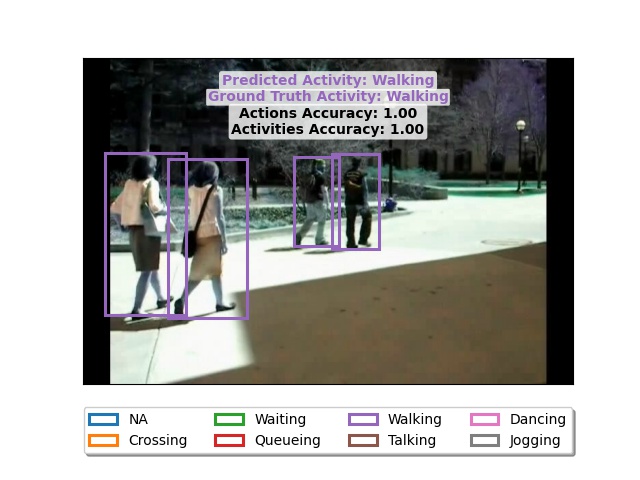}
        \label{fig:2}
    \end{subfigure}
    
     \begin{subfigure}[b]{0.49\columnwidth}  
        \centering 
        \includegraphics[width=\columnwidth]{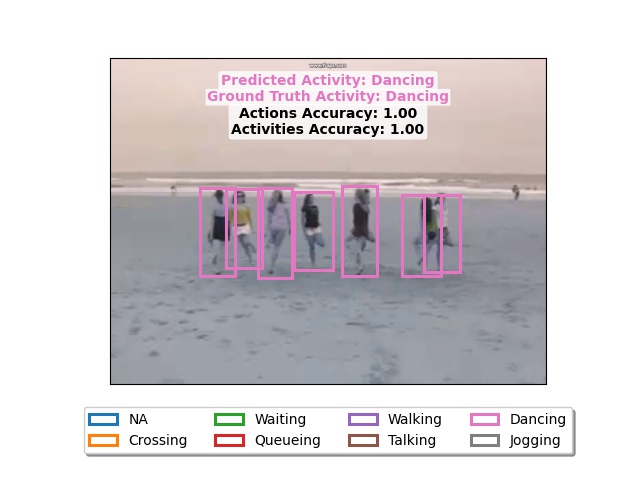}
        \label{fig:2}
    \end{subfigure}
     \begin{subfigure}[b]{0.49\columnwidth}  
        \centering 
        \includegraphics[width=\columnwidth]{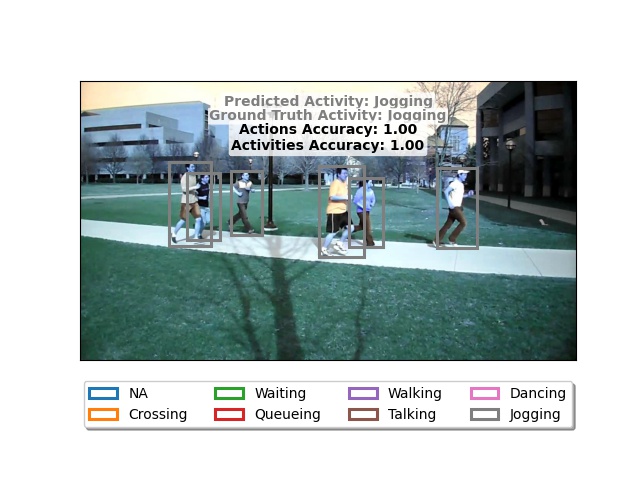}
        \label{fig:2}
    \end{subfigure}
    \caption[]
    {\small Visualization of results on 6 test video clips} 
    \label{fig:3}
\end{figure*}

\end{document}